\documentclass[letterpaper, conference]{IEEEtran}
\IEEEoverridecommandlockouts

\usepackage{times}
\usepackage{epsfig}
\usepackage{graphicx}
\usepackage{amsmath}
\usepackage{amssymb}
\usepackage{tabularx} 
\usepackage{enumitem} 
\usepackage{dirtytalk}
\usepackage{svg}
\usepackage{caption}
\usepackage{comment}
\usepackage{csquotes}
\usepackage{todonotes}

\newcommand\copyrighttext{%
  \footnotesize \textcopyright 2021 IEEE. Personal use of this material is permitted.
  Permission from IEEE must be obtained for all other uses, in any current or future
  media, including reprinting/republishing this material for advertising or promotional
  purposes, creating new collective works, for resale or redistribution to servers or
  lists, or reuse of any copyrighted component of this work in other works.}

\newcommand\copyrightnotice{%
\begin{tikzpicture}[remember picture,overlay]
\node[anchor=south,yshift=10pt] at (current page.south) {\fbox{\parbox{\dimexpr\textwidth-\fboxsep-\fboxrule\relax}{\copyrighttext}}};
\end{tikzpicture}%
}

\begin{document}

\title{Smart Infrastructure: A Research Junction}

\author{\IEEEauthorblockN{Manuel Hetzel\IEEEauthorrefmark{1},
Hannes Reichert\IEEEauthorrefmark{1},
Konrad Doll\IEEEauthorrefmark{1}, 
Bernhard Sick\IEEEauthorrefmark{2}}

\IEEEauthorblockA{\IEEEauthorrefmark{1}University of Applied Sciences Aschaffenburg, Germany\\
Email: \{manuel.hetzel, hannes.reichert, konrad.doll\}@th-ab.de}

\IEEEauthorblockA{\IEEEauthorrefmark{2}University of Kassel, Germany\\
Email: bsick@uni-kassel.de}}

\maketitle

\copyrightnotice

\begin{abstract}
Complex inner-city junctions are among the most critical traffic areas for injury and fatal accidents. The development of highly automated driving (HAD) systems struggles with the complex and hectic everyday life within those areas. Sensor-equipped smart infrastructures, which can communicate and cooperate with vehicles, are essential to enable a holistic scene understanding to resolve occlusions drivers and vehicle perception systems for themselves can not cover. We introduce an intelligent research infrastructure equipped with visual sensor technology, located at a public inner-city junction in Aschaffenburg, Germany. A multiple-view camera system monitors the traffic situation to perceive road users' behavior. Both motorized and non-motorized traffic is considered. The system is used for research in data generation, evaluating new HAD sensors systems, algorithms, and Artificial Intelligence (AI) training strategies using real-, synthetic- and augmented data. In addition, the junction features a highly accurate digital twin. Real-world data can be taken into the digital twin for simulation purposes and synthetic data generation.
\end{abstract}

\section{INTRODUCTION} \label{introduction}
One of the most challenging locations for drivers and HAD systems are inner-city junctions. Extensive traffic density highly restricts the field of view (FOV) of drivers and vehicle-based perception systems. Towards reliable HAD systems, it is mandatory to investigate to what extent these restrictions can be compensated from a vehicle perspective. Intelligent junctions equipped with sensors are already used to cope with these restrictions cooperatively \cite{goldhammer2012koper} \cite{dlr} \cite{conti}. Moreover, one can use them to evaluate the perception capabilities of vehicle-only HAD systems. Carefully matched-up sensor positions, alongside empiric perception models can dissolve almost all occlusions, allowing a seamless scene understanding at complex junctions. Empirical approaches are necessary to satisfy these use cases. These topics are part of the cooperative research project $\textit{AI Data Tooling}$ \cite{kidt}. The project will develop and investigate holistic tools and methods for providing data of different sensor modalities for AI-based functions. The aim is to develop a complete data solution for the training and validation of AI-based automated driving functions by integrating real data, synthetically generated data, and augmented data as a mixture of these two and methods for the efficient handling of this data set. The presented sensor setup facilitates a multi-view perception of traffic participants, with a broad area coverage among the junction. This publication focuses on installation, data flow, and research targets of intelligent infrastructure at a junction.

Moreover, it underlines the possibilities for future research within road safety and training strategies using a mixture of real-, synthetic- and augmented data. The publication is structured as follows: First, we review comparable research junctions in Sec. \ref{state_of_the_art}. Second, we discuss the requirements and analysis for developing the infrastructure perception system in Sec. \ref{requirements_and_analysis}. Next, the system architecture, data recording, and digital twin are described in Sec. \ref{system_description}, followed by challenges and research targets in Sec. \ref{challenges_research_targets}. Finally, we summarize the current status in Sec. \ref{conclusion}.
\section{State of the Art} \label{state_of_the_art}
This chapter references other intelligent junctions used for vulnerable road users' (VRUs) safety and reviews the research work carried out on data acquired by those. Several environmental observing junctions have been proposed. However, the main focus of the majority of research is carried out on high-level traffic flow understanding \cite{survey}. In contrast, some research targets VRU safety topics, which require high-resolution sensing technologies. One of them was introduced in 2012 for the German $\textit{Ko-PER}$ project \cite{goldhammer2012koper}. A combination of laser scanners and gray-scale cameras is used to monitor the traffic of the whole junction in general. In addition, a precise 90-degree stereo camera setup using two gray-scale full HD cameras has been used to detect and predict VRUs behavior crossing the street, focusing on one corner of the junction. Furthermore, this setup is used for the $\textit{DeCoInt}^2$ \cite{decoint} project with the research target of detecting intentions of VRUs based on collective intelligence, focusing on cyclists. The $\textit{DeCoInt}^2$ project covered two major research areas: perception, and motion anticipation, both under the cooperative aspect between static mounted sensors and mobile research vehicles. For motion anticipation, Reitberger et al. provided a cooperative tracking algorithm for cyclists \cite{reitberger2018cooperative}, Bieshaar et al. used Convolutional Neural Networks to detect starting movements of cyclists \cite{Bieshaar}, and Zernetsch et al. developed a probabilistic VRU trajectory forecasting method \cite{zernensch}. Kress et al. used this sensor setup as a reference to evaluate a human keypoint detection model deployed to a mobile research vehicle \cite{kress}. It is worth mentioning that this sensor setup and the knowledge from the $\textit{Ko-PER}$ and $\textit{DeCoInt}^2$ projects was utilized for the development of the novel proposed sensor setup.

A comparable junction is located in Braunschweig, Germany, serving as a field instrument for detecting and assessing traffic behavior. The junction can provide trajectory data of road users, acquired by multi-modal sensor setups. Mono cameras and radar are utilized for the 3D detection of vehicles. For VRU detection, multiple binocular stereo camera setups facing the pedestrian crossings are used. \cite{dlr}

Since 2019 Continental operates two intelligent junctions in public use in Auburn Hills, Michigan. The systems are used to improve traffic flow, reduce pollution, and increase the junction’s safety by communicating hidden dangers to approaching connected vehicles and pedestrians. Camera and radar sensors are used to create an environment model providing information about road users, traffic infrastructure, static objects to connected vehicles using infrastructure-to-everything (I2X) communication \cite{conti}.
\section{REQUIREMENTS AND ANALYSIS} \label{requirements_and_analysis}
The requirements for the original $\textit{Ko-PER}$ junction introduced in 2012 are derived from intensive analysis of accident scenarios which occur at junctions \cite{goldhammer2012koper}. The analysis is based on the German In-Depth Accident Study (GIDAS) database \cite{gidas}. It contains more than 20,000 registered accidents in the area of Hannover and Dresden since the year 1999. According to GIDAS, Goldhammer et al. clustered a total of 29 types of relevant intersection accidents, focusing on pedestrians, into five scenarios, covering 74.8\% of severe and lethal accidents. Accident scenarios are complex due to many different influences like the quickly changing number and variety of road users, complex intersection layout, speed ranges, and different directions from which traffic may approach. 71 \% of all accidents, including pedestrians and 58 \% of all accidents including cyclists in Europe, happen inside urban areas, mostly at intersections and crosswalks \cite{EuropeRoadSafteyReport}. Furthermore, over 60 \% of all fatalities occurring at junctions are of VRUs. We used these results as a baseline for our requirement analysis. According to GIDAS inter-vehicle and vehicle to pedestrian accident rates at junctions are continuously decreasing, from 43\% in 2010 to 33\% in 2020. This development illustrates the impact of improved driver assistance systems. In contrast, vehicle to bicycle, vehicle to "others", and "inter-VRU" accidents increased, from 42\% in 1999 to 67\% in 2020. The group "others" represent road users like electric bikes and scooters, still belonging to VRUs. Based upon this development and in contrast to $\textit{Ko-PER}$ or $\textit{DeCoInt}^2$, we are focusing on VRUs in general (i.e. pedestrians, cyclists and "others").

For accurate VRU detection and motion anticipation, high-resolution image-based sensors are essential. For determining positions of VRUs, seamless stereoscopic coverage of the VRU relevant areas at the junction is necessary. To further minimize occlusions, the sensors need to be mounted several meters above street level. 

In addition, seasonal and weather conditions might be challenging for image-based sensors. Thus, we need to monitor the weather and analyze the impact of weather effects on image-based perception. 

For AI-based perception and prediction methods, so-called corner cases that are critical and rarely occurring are essential. Corner cases can efficiently be generated using simulations. A precise environmental model, including textures and material properties, is required to simulate situations and scenarios at the junction as accurately as possible. 

A suitable image acquisition frame rate is required to deal with a wide range of possible velocities for both motorized and non-motorized road users. We aim to maximize the frame rate while keeping both amount of data and computational load manageable.
\section{SYSTEM DESCRIPTION} \label{system_description}

\begin{figure}[h!]
    \centering
    \includegraphics[clip,width=\columnwidth]{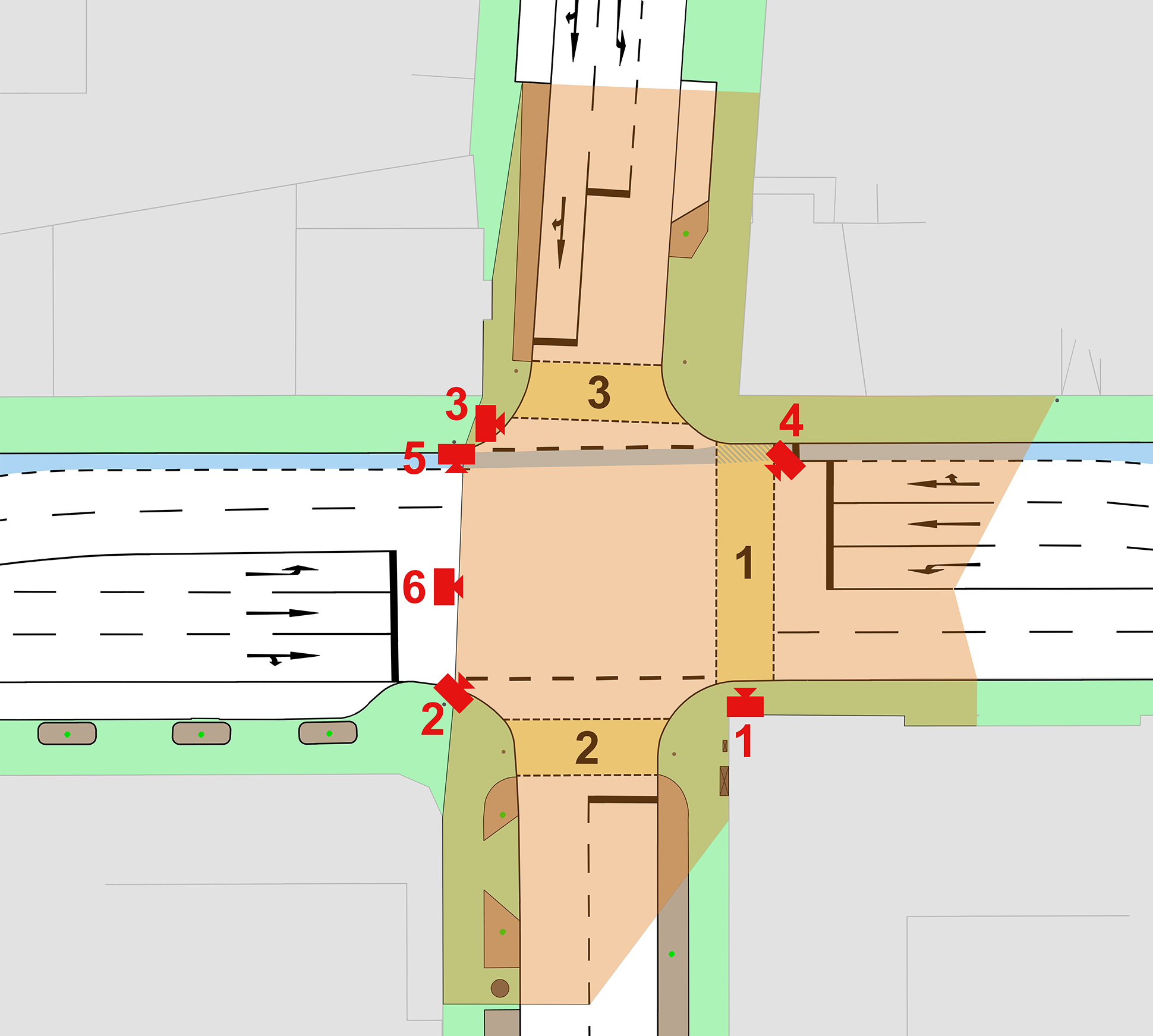}
    \caption{Illustration of the junctions topology, including an overview of all camera positions (red) with full stereo coverage in the highlighted area (orange).}
    \renewcommand{\the}{}
    \label{fig:stereo_fov}
\end{figure}


The junction consists of the main road with five lanes and a daily traffic volume of 30,000 vehicles. Fig. \ref{fig:stereo_fov} gives a schematic overview of the junction's topology. The main road has two straight-ahead lanes at the junction area and a separate left-turn lane for each direction. The more minor approaches have one lane per direction and a left-turn lane on one side. There are three traffic light-controlled crosswalks and a bicycle lane along the main road, which VRUs highly frequent, due to proximity to a university. The four corners of the junction show different occlusions by roadside structures and parking cars. Thus, the road users' FOV in many common traffic scenarios is limited.

\subsection{Sensor Setup}

In advance of installing the multi-camera network setup, the FOV of all sensors was simulated for different positions and alignments. The sensor setup was adjusted to achieve a best-case stereo coverage of the inner junction area, including the three pedestrian crosswalks, a bicycle lane, and a sensor coverage up to 100 meters into the junction approaches. Elevated mounting positions, up to eight meters, alongside overlapping FOVs, dissolve occlusions that may appear. All sensors are mounted to existing infrastructural light and signal poles. The optical sensor network consists of six identical color CMOS cameras with an ultra-high-resolution of 4096x2160 pixels. Each camera is equipped with a 71-degree horizontal aperture angle lens,  operates at a fixed 25 Hz acquisition rate, and uses a five GigE interface to submit its acquisition data. Every camera is placed within a weather-resistance case, including a temperature-controlled heating and cooling system for all-season commitment. The transmission of sensor data is done via five and ten Gigabit Ethernet uplinks using fiber and copper cables with lengths of 80 meters. The cameras are aligned to achieve multiple 45- and 90-degrees stereo setups. In total, seven stereo systems are used. The complete stereo FOV is illustrated in Fig. \ref{fig:stereo_fov}. Cameras 1-3 focus on the section between pedestrian crosswalks one and three. Cameras 4-6 focus on the section between pedestrian crosswalks two and three. In addition, cameras one, five, and six covers the three connected approaches corresponding to the pedestrian crosswalks, whereas cameras two, three, and four each cover two approaches by reduced attention. 

\begin{figure}[h!]
    \centering
    \includegraphics[width=\columnwidth]{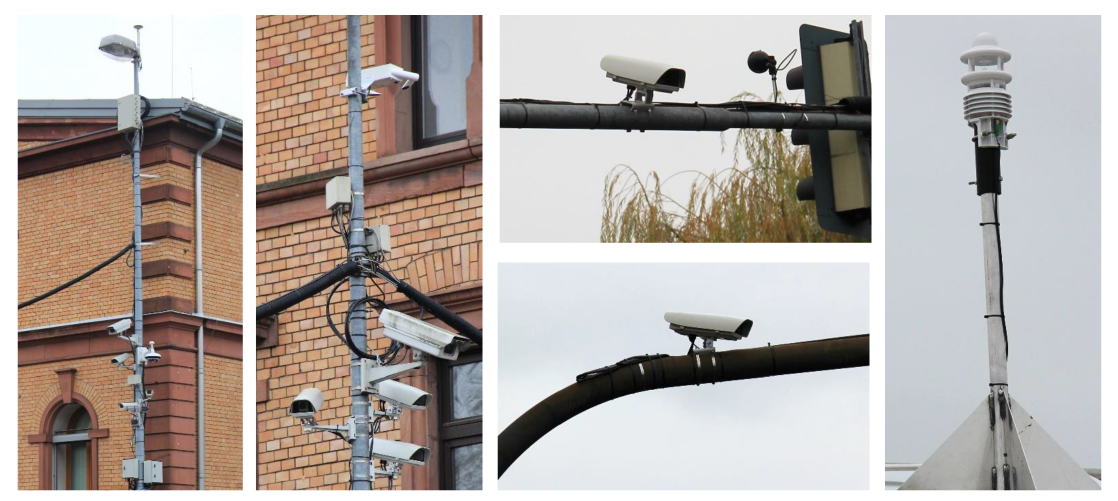}
    \caption{Different mounting positions of sensors.}
    \label{fig:sensors_overview}
\end{figure}

In addition to the image-based sensor setup, a meteorological station is used to provide weather information as supplementary context. The station consists of two sensors placed in proximity to the junction. One sensor is responsible for measuring environmental parameters and is located on a building's roof, next to the junction. A second sensor is used to track the current visibility at the junction. Thus it is placed at a light pole next to the junction center ten meters above road level. Both sensors are shown in Fig. \ref{fig:sensors_overview}.

\subsection{System Architecture}

\begin{figure}[h!]
    \centering
    \includegraphics[clip,width=\columnwidth]{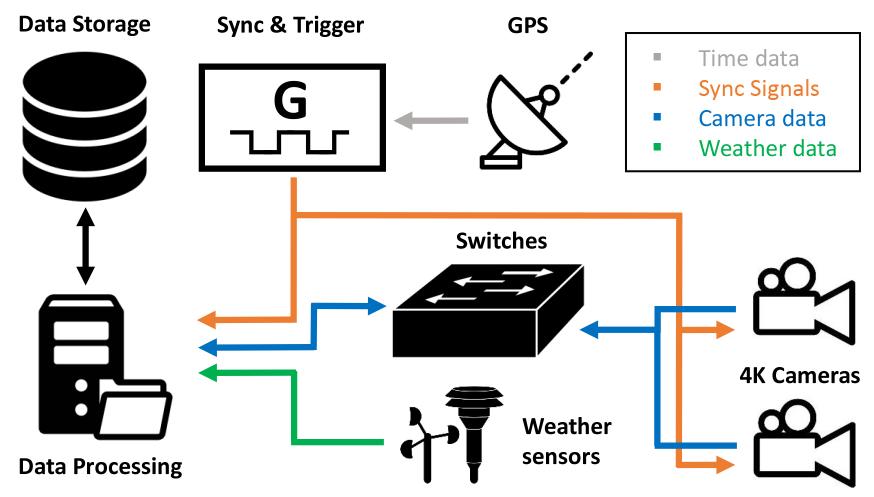}
    \caption{Schematic overview of the system architecture.}
    \label{fig:architectural_overview}
\end{figure}

To handle the amount of data six ultra-high-resolution color cameras provide, a custom build hardware- and software stack is required to maintain real-time data recording and processing in a single system. By using 25 Hz, each camera transmits 1.77 Gigabits of data, resulting in a total amount of 10.62 Gigabits of processing data from all cameras. Fig. \ref{fig:architectural_overview} illustrates the complete schematic system architecture with the sensor setup described in the previous section. The data processing system consists of a 64-Core processor with 256 GB of RAM, three GPUs, four TB of PCIe 4.0 NVMe storage, and ten high-speed Ethernet ports. The GPUs are necessary for high-speed image data encoding. On the CPU side, our software stack makes heavy use of multithreading for simultaneous data handling. In addition to simple sensor data recording, the system can serve real-time demonstrations, including 3D perception and VRU motion anticipation.
For highly precise synchronization, the cameras are triggered by GPS timestamps. Simultaneously, dedicated UTC timestamps are sent to the data processing system associated with the sensor data by CAN-Bus. A storage server with 576 TB of capacity is connected to the data processing system to store more extensive data sets. The Robot operation system (ROS) manages the complete data handling within the processing system. It enables the establishment of a flexible node-based data processing pipeline using a publisher-subscriber architecture.

\subsection{Data Recording} \label{encoding}

As described in the previous subsection, the cameras provide a data stream of more than 10 Gigabits. To ensure a continuous and seamless data recording, GPU fixed encoding hardware functionalities are used. Each GPU can processes two camera streams simultaneously. The fixed-function unit encodes the camera's raw data into the lossless compressed H.264/MPEG-4 AVC format \cite{h264}. Using the H.264 compression algorithm, we can reduce the amount of data by a factor of eight to ten on average, depending on the current junction traffic volume. The recording node itself can subscribe to a user-defined number of cameras. GPU resources are automatically managed. Within a recording session, the camera images are uploaded into the GPU memory and passed to the hardware-accelerated encoding unit. Afterward, the resulting H.264 binary stream is stored on disk. A synchronization file is created to keep the UTC timestamps.

\subsection{Meta Data}

Besides the raw data recording capabilities, the system performs several post-processing tasks to create a wide range of metadata for additional research topics. We are using Detectron2 for state-of-the-art object detection, segmentation and human pose extraction \cite{detectron2}. In addition, triangulation is applied for each stereo-system to determine 3D coordinates for all detected VRUs. By merging all seven stereo-systems, we maintain a complete 3D perception of the junction's critical areas, as illustrated in Fig. \ref{fig:stereo_fov}. This allows us to track objects in real-world coordinates. Furthermore, the system can estimate 3D human body poses, as introduced by Open Pose \cite{openpose}. The H.264 encoding mentioned in subsection \ref{encoding} can be used to extract optical flow, which is a powerful input feature for the task of human motion anticipation in general, as shown by Carreira and Zisserman \cite{deepmind} and for VRU motion anticipation in particular, as shown by Zernetsch \cite{ZSK21}.

\subsection{Digital Twin}

A digital clone of the junction is created by a combination of photogrammetry and road-level laser scans. A drone carrying a specific high-resolution camera is used to receive the visual scan of the junction, supported by measurement vehicles to scan the complete ground area and facades. Both methods achieve a highly accurate digital model with a better than 1 cm textural resolution and a 3 cm or better structural resolution. The model can be utilized in simulation environments, increasing the junction's research capabilities in synthetic data creation and applications.
\section{RESEARCH TARGETS AND CHALLENGES } \label{challenges_research_targets}
Based on the data collection capability enabled by the junction, we want to envision future research directions. Unlike ground-based vehicles, our system can sense objects of interest without occlusion. For HAD vehicles, perception systems have to be safe and reliable. Due to the high-quality data and its multiple perspectives, our sensor setup can serve as a reference system for evaluating and safeguarding vehicular HAD systems. That can be done in the form of a test site or by utilizing the data to train vehicular HAD systems by transferring labels from the reference system.
Furthermore, the digital twin enables data generation and evaluation by simulation.  We want to collect data, analyze critical scenarios, define evaluation metrics, develop methods for meta-data acquisition, and provide a simulation environment for the digital twin. 

The meta-data provides an interface between the real world and its digital twin, e. g. human body poses and trajectories can be used to animate pedestrians and cyclists within a simulation environment. Both synthetic and real data allow us to analyze the quality requirements of synthetic data and how well the mixture of real and synthetic data works for use cases like object detection and motion anticipation. Using synthetic data alongside real data is promising, as it allows us to teach HAD systems even those scenarios that rarely occur.

For an accurate sensor system, several challenges must still be overcome. One is a continuous calibration of the cameras to compensate for drifts in position and orientation caused by temperature and mechanical vibrations. The second is maintaining suitable perception models within the presented system regarding changes in seasons, traffic patterns, and urban mobility. For example, electric bicycles change the way human car drivers interact with cyclists within a few years.
\section{CONCLUSIONS} \label{conclusion}
This paper presented a sensor network operating at a complex public junction as part of the $\textit{AI Data Tooling}$ project, including time-associated data acquisition and data processing. Moreover, we highlighted the research capabilities of our system and defined future research targets. The dedicated setup of ultra-high-resolution cameras enables a highly accurate perception of all road users within the inner junction area. It will be used as a reference for vehicle sensor evaluation and simulation, data creation, and analyzing AI training strategies using real- synthetic- and augmented data.
\section{\large Acknowledgment}

This work results from the KI Data Tooling supported by the Federal Ministry for Economic Affairs and Energy (BMWi), grant numbers, 19A20001L and 19A20001O.


{\small
\bibliographystyle{ieeetr}
\bibliography{egbib}
}

\end{document}